\documentclass[11pt,twoside]{article}

\usepackage[utf8]{inputenc}
\usepackage{graphicx}
\usepackage{amsmath,amssymb,amsfonts,bm,mathtools}
\usepackage{hyperref}
\usepackage{natbib}
\usepackage{fancyhdr}
\usepackage{geometry}
\usepackage{titlesec}
\usepackage{lipsum}
\usepackage{orcidlink}
\usepackage{booktabs}
\usepackage{placeins}
\usepackage{float}
\usepackage{array}
\usepackage[ruled,vlined,linesnumbered]{algorithm2e}
\usepackage{microtype}
\usepackage{dblfloatfix}

\setcitestyle{authoryear}
\hypersetup{
colorlinks=true,
linkcolor=blue,
filecolor=blue,
citecolor=blue,
urlcolor=black
}

\fancypagestyle{plain}{
\fancyhf{}

}

\title{
\vspace{-1.5em}
\hrule height 1.5pt
\vspace{0.8em}
StrEBM: A Structured Latent Energy-Based Model for Blind Source Separation
\vspace{0.8em}
\hrule height 1.5pt
\vspace{1em}
}

\author{%
\begin{minipage}[t]{.48\textwidth}\centering\small
  \textbf{Yuan-Hao Wei\orcidlink{0000-0001-9439-0780}}\\
  \texttt{Yuan-Hao.Wei@outlook.com; yuan-hao.wei@connect.polyu.hk; yuanhao.wei1993@gmail.com}
\end{minipage}
}

\date{}

\begin{document}

\twocolumn[
\maketitle
\thispagestyle{plain}

\begin{abstract}
    This paper proposes StrEBM, a structured latent energy-based model for source-wise structured representation learning. The framework is motivated by a broader goal of promoting identifiable and decoupled latent organization by assigning different latent dimensions their own learnable structural biases, rather than constraining the entire latent representation with a single shared energy. In this sense, blind source separation is adopted here as a concrete and verifiable testbed, through which the evolution of latent dimensions toward distinct underlying components can be directly examined. In the proposed framework, latent trajectories are optimized directly together with an observation-generation map and source-wise structural parameters. Each latent dimension is associated with its own energy-based formulation, allowing different latent components to gradually evolve toward distinct source-like roles during training. In the present study, this source-wise energy design is instantiated using Gaussian-process-inspired energies with learnable length-scales, but the framework itself is not restricted to Gaussian processes and is intended as a more general structured latent EBM formulation. Experiments on synthetic multichannel signals under linear and nonlinear mixing settings show that the proposed model can recover source components effectively, providing an initial empirical validation of the framework. At the same time, the study reveals important optimization characteristics, including slow late-stage convergence and reduced stability under nonlinear observation mappings. These findings not only clarify the practical behavior of the current GP-based instantiation, but also establish a basis for future investigation of richer source-wise energy families and more robust nonlinear optimization strategies.
\end{abstract}

\vspace{0.8em}
\vspace{1.5em}
]

\section{Introduction}
Blind source separation (BSS) is concerned with recovering latent source components from observed mixtures without direct access to the underlying sources. As a classical problem in signal processing and statistical learning, it has long been studied through the framework of independent component analysis (ICA), where separation is achieved by exploiting statistical independence or non-Gaussianity among latent factors (\cite{jutten1991blind,comon1994independent,bell1995information,hyvarinen2000independent}). In the linear case, this viewpoint has led to a rich body of theory and highly influential algorithms. However, once the observation process becomes nonlinear, or once one moves beyond purely iid assumptions toward temporally or structurally organized latent factors, the problem becomes substantially more difficult. In such settings, the challenge is no longer merely to reconstruct observations, but to induce latent variables that evolve toward distinct, interpretable, and potentially identifiable roles.

This broader perspective connects BSS to a wider line of work on disentangled and structured latent representation learning. A central ambition in that literature is to recover latent factors that correspond, at least approximately, to distinct explanatory causes of the data. Yet it is now well understood that unsupervised disentanglement cannot generally be expected to emerge without suitable inductive biases or auxiliary information (\cite{locatello2019challenging}). This observation is especially relevant for source separation and nonlinear ICA, where identifiability is a fundamental issue. Recent progress on nonlinear ICA has shown that identifiability can be recovered by exploiting additional structure, such as temporal nonstationarity, auxiliary variables, or other side information (\cite{hyvarinen2016unsupervised,hyvarinen2019nonlinear,khemakhem2020variational}). These results suggest an important principle: latent factors become more recoverable when different dimensions are subjected to sufficiently informative and non-exchangeable structural biases.

In parallel, deep latent-variable models have provided flexible mechanisms for learning nonlinear generative mappings from latent representations to observations. Variational autoencoders (VAEs) made such models practical by combining amortized inference with stochastic gradient optimization (\cite{kingma2014auto}). A large subsequent literature has sought to modify the latent objective so as to encourage factorization or disentanglement, for example through weighted KL regularization or total-correlation penalties (\cite{higgins2017beta,chen2018isolating}). While these methods often improve axis-wise factorization, they typically still rely on relatively simple latent priors and are not, by themselves, sufficient to guarantee identifiability in the absence of stronger inductive assumptions (\cite{locatello2019challenging}). More importantly for the present paper, many of these approaches regularize the latent representation in a largely shared or global manner, rather than explicitly endowing each latent dimension with its own learnable structural identity.

To address this limitation, another line of research has incorporated explicit probabilistic structure into latent-variable models. Structured variational autoencoders and related models combine neural observation mappings with graphical-model or state-space structure in the latent variables, thereby introducing temporal dependence, dynamical constraints, or other forms of latent organization (\cite{johnson2016composing,karl2017deep,zhao2023revisiting}). Gaussian-process-based latent modeling provides a particularly clear example of this idea. By replacing the standard iid latent prior with GP-induced covariance structure, models such as GPPVAE and GP-VAE demonstrate that structured temporal dependence can improve representation quality, interpretability, and downstream inference for sequential or correlated data (\cite{casale2018gaussian,fortuin2020gp}). These developments reinforce the broader message that latent structure is not merely a regularization detail, but can play a decisive role in shaping what the model is able to recover.

At the same time, energy-based models (EBMs) provide a complementary and highly flexible framework for latent modeling. Instead of requiring a normalized probability model with a fixed tractable prior, EBMs associate low energies with preferred configurations and higher energies with less plausible ones (\cite{lecun2006tutorial}). This flexibility makes them attractive for representing rich dependencies and structured constraints, but it also brings training and optimization challenges, especially in deep settings (\cite{yu2020training}). Despite the generality of the EBM viewpoint, comparatively less attention has been devoted to using source-wise or dimension-wise energy design as the main mechanism for organizing latent factors into distinct roles. In particular, the idea that each latent dimension should possess its own learnable energy geometry, so that the latent coordinates may progressively differentiate during training, remains insufficiently explored.

This gap is especially notable when viewed from the perspective of identifiable and decoupled latent organization. Classical ICA imposes independence as a global statistical principle (\cite{comon1994independent,hyvarinen2000independent}), while more recent nonlinear ICA methods recover identifiability by introducing auxiliary structure shared across the learning problem (\cite{hyvarinen2016unsupervised,hyvarinen2019nonlinear,khemakhem2020variational}). Likewise, disentanglement-oriented VAEs often encourage factorization through global penalties (\cite{higgins2017beta,chen2018isolating}), and structured latent-variable models typically specify one structured prior family over the latent process as a whole (\cite{johnson2016composing,karl2017deep,casale2018gaussian,fortuin2020gp}). What is still less studied is a source-wise structured latent formulation in which different latent dimensions are directly assigned different learnable structural energies, so that separation or decoupling emerges through the joint adaptation of latent trajectories, observation mappings, and source-specific structural parameters.

Motivated by this gap, this paper proposes \emph{StrEBM}, a structured latent energy-based model for source-wise structured representation learning. The key idea is to move away from a single shared latent regularizer and instead assign each latent dimension its own learnable energy-based formulation. Under this design, the latent trajectories are optimized directly, together with the observation-generation map and the source-wise structural parameters. Thus, the latent dimensions are not treated as interchangeable coordinates under one common prior, but as components that may gradually evolve toward different structural roles during training. Blind source separation is adopted here as a concrete and verifiable testbed for this broader idea, because in that setting the learned latent dimensions can be directly compared with underlying source components.

In the present study, the source-wise energy is instantiated using Gaussian-process-inspired energies. This GP-based choice is not intended to define the general framework exhaustively, but rather to provide one natural realization for indexed one-dimensional latent trajectories. In this realization, each latent dimension has its own learnable length-scale, which determines the covariance geometry underlying its structural energy. Consequently, different latent dimensions are encouraged to settle into different smoothness or correlation regimes, rather than sharing one common latent bias. This design is conceptually related to the role of structured priors in temporal latent-variable models (\cite{casale2018gaussian,fortuin2020gp}), but differs in two important respects: first, the formulation is explicitly source-wise rather than globally latent-structured; second, the latent trajectories themselves are optimized directly in an EBM-style framework instead of being inferred solely through an encoder-based variational posterior.

This source-wise design is also motivated by a broader question beyond BSS itself. If identifiable or decoupled latent organization requires inductive bias (\cite{locatello2019challenging}), then one possible strategy is to impose not just \emph{some} latent structure, but \emph{different} learnable structures on different latent dimensions. In that sense, the present paper studies BSS as an experimentally transparent setting in which the latent dimensions can be inspected and matched to known underlying components, while the deeper methodological interest lies in structured latent organization through source-wise energies. The hope is that such a formulation may eventually extend beyond source separation to more general problems of interpretable representation learning, identifiable latent factorization, and source-wise modeling of structured data.

The experiments in this paper provide an initial empirical study of this idea under both linear and nonlinear mixing settings. The results show that the proposed framework can effectively recover source components, supporting the basic feasibility of source-wise structured latent EBM learning in this setting. At the same time, the experiments also reveal two important practical phenomena. First, although source-wise structural energy alone can already drive separation in the linear case, convergence may become very slow in the later stage, which motivates the introduction of a weak auxiliary separation regularizer. Second, under nonlinear observation mappings, the optimization becomes less stable, suggesting that the interaction among flexible generators, directly optimized latent trajectories, and source-wise structural energies remains challenging. These observations are important not only for assessing the present GP-based instantiation, but also for motivating future work on richer energy families, improved optimization strategies, and more robust nonlinear source-wise latent modeling.

The main contribution of this paper is therefore not merely a new GP-based BSS model, but a first step toward a broader source-wise structured latent EBM framework. By assigning each latent dimension its own learnable structured energy and by directly optimizing the latent trajectories together with the observation map, the proposed approach offers a different route to latent decoupling: separation is not imposed through explicit supervision, nor does it arise solely from a shared latent prior, but instead emerges through the progressive differentiation of latent roles under source-wise structural energies. In this sense, the present work contributes both a concrete model and an initial experimental analysis, while laying the groundwork for future investigations into source-wise energy design for identifiable and interpretable latent representation learning.

\section{Methodology}

\subsection{Problem Setup and Latent Representation}

The proposed framework is motivated by a broader goal in structured latent modeling: to endow different latent dimensions with their own learnable structural biases, so that they may evolve into distinct, interpretable, and potentially more identifiable latent factors during training. Rather than treating the latent representation as a single undifferentiated vector governed by one shared prior, the present formulation assigns an individual structured energy to each latent dimension. In principle, such a mechanism is relevant to disentangled representation learning, source-wise interpretability, and identifiable latent organization. In the present work, this idea is instantiated in a comparatively simple and verifiable setting, namely latent factor recovery from multichannel observations, where the learned latent dimensions can be directly examined against underlying components. In this sense, the present study may also be viewed as a blind source separation problem, in which the latent dimensions are encouraged to align with underlying source components from multichannel observations while remaining embedded in a more general source-wise structured latent modeling framework.

Let $\mathbf{Y}\in\mathbb{R}^{T\times m}$ denote the observed multichannel signal, where $T$ is the number of indexed samples and $m$ is the number of observed channels. The latent representation is written as
\begin{equation}
\mathbf{S}=[\mathbf{s}_1,\mathbf{s}_2,\dots,\mathbf{s}_n]\in\mathbb{R}^{T\times n},
\end{equation}
where each column $\mathbf{s}_j\in\mathbb{R}^{T}$ is one structured latent component defined over an ordered index set. In the current instantiation, these latent dimensions are expected to align with underlying source components, but the formulation itself is introduced from the more general viewpoint that different latent dimensions should be allowed to develop different structural roles. In the present implementation, the ordered index is instantiated by the normalized sample index, which may be interpreted, for example, as a normalized time index in the current experiments.

Unlike encoder-based variational formulations, the present method directly treats the latent matrix as an optimization variable. Specifically, the initial latent components are randomly initialized as
\begin{equation}
\mathbf{S}^{(0)}=\sigma_{\mathrm{init}}\boldsymbol{\Xi},
\qquad
\Xi_{ij}\sim\mathcal{N}(0,1),
\label{eq:init_s}
\end{equation}
where $\sigma_{\mathrm{init}}>0$ is the initialization scale. Thus, at the beginning of training, each latent dimension is simply a random indexed trajectory rather than the output of an encoder or an explicit structured prior sampler. During optimization, these trajectories are gradually shaped by the observation-matching term, source-wise structural energy, and a weak separation regularizer. This matches the implementation in the code, where the latent trajectories are initialized by Gaussian noise and then optimized directly as trainable parameters.

Once the structured latent representation has been introduced, the next question is how these latent components give rise to the observed multichannel signal. This leads to the observation-generation stage of the framework.

\subsection{Observation Mapping and Reconstruction Term}

The next step of the framework specifies how the structured latent representation gives rise to the observed multichannel signal. At a general level, this stage describes the mapping from structured latent components to the observation domain, and may therefore be viewed as the generative process from the latent space to the observed space. In the particular blind source separation interpretation adopted in this paper, the same generative mapping is instantiated more concretely as a mixing mechanism that transforms latent source-like components into observed mixtures.

For each index location $i\in\{1,\dots,T\}$, let $\mathbf{s}(i)\in\mathbb{R}^{n}$ denote the $i$-th row of $\mathbf{S}$. The latent representation is mapped to the observation space according to
\begin{equation}
\widehat{\mathbf{y}}(i)=g_{\theta}\big(\mathbf{s}(i)\big),
\end{equation}
where $g_{\theta}:\mathbb{R}^{n}\rightarrow\mathbb{R}^{m}$ is a parametric observation map. Stacking all index locations yields
\begin{equation}
\widehat{\mathbf{Y}}=g_{\theta}(\mathbf{S})\in\mathbb{R}^{T\times m}.
\end{equation}

Accordingly, $g_\theta$ serves in this work as a general observation-domain generator for structured latent variables. Under the present blind source separation viewpoint, however, it is more specifically interpreted as a mixing map from latent source-like components to observed mixtures. This terminology is especially natural in the current setting because the latent dimensions are expected to evolve toward underlying source components, while the observation map explains how these components combine to form the measured channels.

When no hidden layer is used, the observation map reduces to a linear form,
\begin{equation}
\widehat{\mathbf{y}}(i)=\mathbf{W}\mathbf{s}(i)+\mathbf{b},
\label{eq:linear_mixer}
\end{equation}
with $\mathbf{W}\in\mathbb{R}^{m\times n}$ and $\mathbf{b}\in\mathbb{R}^{m}$. In this case, the observation map coincides with a linear mixer. When hidden layers are introduced, $g_\theta(\cdot)$ becomes a nonlinear multilayer perceptron applied row-wise, so that the same framework also admits nonlinear mixing relationships. Therefore, the present formulation accommodates both linear and nonlinear observation-generation mechanisms within a unified structured latent modeling framework.

The observation-matching term used in the implementation can be interpreted as the negative log-likelihood of an isotropic Gaussian observation model,
\begin{equation}
p(\mathbf{Y}\mid \mathbf{S},\theta)
\propto
\exp\!\left(
-\frac{1}{2\nu_y}
\|\mathbf{Y}-g_\theta(\mathbf{S})\|_F^2
\right),
\label{eq:obs_model}
\end{equation}
where $\nu_y>0$ controls the relative weight of data fidelity (\cite{higgins2017beta,wei2024innovative}). The practical loss is written as
\begin{equation}
\mathcal{L}_{\mathrm{obs}}
=
\frac{1}{2\nu_y}\cdot
\left\|
\mathbf{Y}-\widehat{\mathbf{Y}}
\right\|_F^2.
\label{eq:Lobs}
\end{equation}

From the first epoch onward, the current latent matrix $\mathbf{S}$ is passed through the observation map to produce $\widehat{\mathbf{Y}}$, after which the discrepancy between $\mathbf{Y}$ and $\widehat{\mathbf{Y}}$ drives the latent trajectories toward a representation that better explains the observed multichannel signal. Under the source separation interpretation adopted here, this means that the latent dimensions are progressively encouraged to organize into source-like components whose mixing through $g_\theta$ reproduces the observed mixtures, even though no direct source supervision is provided.

Observation matching alone, however, does not determine how different latent dimensions should organize internally. To induce distinct structural roles across latent components, an additional source-wise structural mechanism is introduced.

\subsection{Source-Wise Structural Energy and the GP Instantiation}

To impose distinct structure on different latent dimensions, the framework introduces a source-wise structural energy for each latent trajectory. At the general level, let
\begin{equation}
E_j(\mathbf{s}_j;\phi_j)
\end{equation}
denote the structured energy assigned to the $j$-th latent dimension, where $\phi_j$ represents the corresponding source-wise structural parameters. This leads to a factorized latent energy model of the form
\begin{equation}
p(\mathbf{S})
\propto
\prod_{j=1}^{n}
\exp\!\big(-E_j(\mathbf{s}_j;\phi_j)\big).
\label{eq:general_structured_prior}
\end{equation}

Equation \eqref{eq:general_structured_prior} expresses the high-level design principle of the present framework: different latent dimensions need not share one common structural prior, but may instead be equipped with their own learnable structural energies. The specific choice of energy is not conceptually restricted to one particular family. In the present paper, a Gaussian-process-inspired realization is adopted as a concrete and natural example for indexed one-dimensional latent components.

Let
\begin{equation}
\mathbf{u}=[u_1,\dots,u_T]^\top
\end{equation}
denote the ordered index associated with the $T$ samples. In the present implementation, $\mathbf{u}$ is given by the normalized sample index, which in the current experiments may be viewed as a normalized time index. For the $j$-th latent dimension, the GP-inspired realization defines an RBF covariance matrix
\begin{equation}
\mathbf{K}_j(i,r)
=
\sigma_f^2
\exp\!\left(
-\frac{(u_i-u_r)^2}{2\ell_j^2}
\right)
+
\epsilon\,\delta_{ir},
\label{eq:Kk}
\end{equation}
where $\ell_j>0$ is the learnable length-scale associated with the $j$-th dimension, $\sigma_f^2$ is a fixed kernel amplitude, $\epsilon>0$ is a small jitter term, and $\delta_{ir}$ is the Kronecker delta.

Each source-wise length-scale is parameterized as
\begin{equation}
\ell_j=\exp(\eta_j),
\label{eq:ellk}
\end{equation}
where $\eta_j\in\mathbb{R}$ is unconstrained. This ensures positivity while allowing different latent dimensions to develop different correlation scales during training.

Under the present GP-inspired choice, the $j$-th structured energy is induced from the Gaussian density
\begin{equation}
\mathbf{s}_j\sim\mathcal{N}(\mathbf{0},\mathbf{K}_j),
\end{equation}
whose density is
\begin{equation}
p(\mathbf{s}_j)
=
\frac{1}{(2\pi)^{T/2}|\mathbf{K}_j|^{1/2}}
\exp\!\left(
-\frac{1}{2}\mathbf{s}_j^\top \mathbf{K}_j^{-1}\mathbf{s}_j
\right).
\end{equation}
Taking the negative logarithm and omitting constants independent of the optimization variables gives the GP-inspired source-wise energy
\begin{equation}
E_j^{\mathrm{GP}}(\mathbf{s}_j;\ell_j)
=
\frac{1}{2}\mathbf{s}_j^\top\mathbf{K}_j^{-1}\mathbf{s}_j
+
\frac{1}{2}\log|\mathbf{K}_j|.
\label{eq:Ek_gp}
\end{equation}
Summing over all latent dimensions yields the structured energy used in this paper,
\begin{equation}
\begin{aligned}
\mathcal{L}_{\mathrm{str}}
&=
\sum_{j=1}^{n}E_j^{\mathrm{GP}}(\mathbf{s}_j;\ell_j) \\
&=
\sum_{j=1}^{n}
\left(
\frac{1}{2}\mathbf{s}_j^\top\mathbf{K}_j^{-1}\mathbf{s}_j
+
\frac{1}{2}\log|\mathbf{K}_j|
\right).
\end{aligned}
\label{eq:Lstr_gp}
\end{equation}

Equation \eqref{eq:Lstr_gp} shows that the latent trajectories are not generated by directly sampling from Gaussian processes. Instead, they are regularized by a source-wise GP-induced energy geometry over the chosen index set. Starting from the random initialization in \eqref{eq:init_s}, each latent dimension is gradually pulled toward a trajectory that attains low energy under its own covariance structure while still remaining useful for explaining the observations. In the present study, this mechanism encourages different latent dimensions to align with different underlying components; more generally, it serves as one concrete realization of a broader source-wise structured latent modeling principle. Although the current implementation uses an ordered one-dimensional index, the same viewpoint is not conceptually restricted to physical time and may also be interpreted with respect to other ordered sample domains when appropriate.

For numerical stability, the covariance matrix is factorized as
\begin{equation}
\mathbf{K}_j=\mathbf{L}_j\mathbf{L}_j^\top,
\end{equation}
where $\mathbf{L}_j$ is the Cholesky factor. The quadratic term is then evaluated by solving triangular linear systems rather than explicitly inverting $\mathbf{K}_j$, and the log-determinant is computed as
\begin{equation}
\log|\mathbf{K}_j|
=
2\sum_{i=1}^{T}\log (\mathbf{L}_j)_{ii}.
\label{eq:logdet_chol}
\end{equation}
Accordingly, the GP-inspired energy in practice is evaluated as
\begin{equation}
E_j^{\mathrm{GP}}(\mathbf{s}_j;\ell_j)
=
\frac{1}{2}\mathbf{s}_j^\top\mathbf{K}_j^{-1}\mathbf{s}_j
+
\sum_{i=1}^{T}\log (\mathbf{L}_j)_{ii},
\label{eq:Ek_gp_chol}
\end{equation}
which is algebraically equivalent to \eqref{eq:Ek_gp}.

While the source-wise structural energies encourage different latent dimensions to settle into different structural configurations, they do not by themselves explicitly suppress residual overlap among latent components during optimization. This motivates the introduction of a weak auxiliary separation regularizer.

\subsection{Separation Regularization}

In the broader source-wise structured latent modeling viewpoint adopted in this work, an explicit separation penalty is not conceptually indispensable. In several of the author's previous studies, the latent dimensions were already able to evolve toward different source-like roles under source-wise structured designs alone, without requiring an additional explicit separation term. In that sense, the primary driving mechanism of separation in the present framework still comes from the combination of observation matching and source-wise structural regularization, rather than from an externally imposed decorrelation constraint.

Nevertheless, the case study considered here also reveals a practical optimization issue. When only the observation term and the source-wise structural energy are used, the model can already reconstruct the observations reasonably well, but the latent trajectories may still remain partially correlated or weakly redundant for a long stage of training. As a result, the later phase of optimization may enter a relatively flat region in which reconstruction is already satisfactory while further refinement toward cleaner source-wise decoupling proceeds only very slowly. Therefore, in the present implementation, a weak separation regularizer is introduced not as the principal source of identifiability, but as an auxiliary term to suppress redundant latent overlap and to accelerate the late-stage convergence toward more distinct source-wise representations.

Let the centered and normalized latent matrix be
\begin{equation}
\widetilde{\mathbf{S}}
=
[\widetilde{\mathbf{s}}_1,\dots,\widetilde{\mathbf{s}}_n],
\end{equation}
where
\begin{equation}
\widetilde{\mathbf{s}}_j
=
\frac{\mathbf{s}_j-\bar{s}_j\mathbf{1}}
{\sqrt{\frac{1}{T}\|\mathbf{s}_j-\bar{s}_j\mathbf{1}\|_2^2+\varepsilon_s}},
\end{equation}
with $\bar{s}_j$ denoting the empirical mean of the $j$-th latent dimension and $\varepsilon_s>0$ a small constant. The empirical correlation matrix of the normalized latent representation is
\begin{equation}
\mathbf{C}
=
\frac{1}{T}\widetilde{\mathbf{S}}^\top\widetilde{\mathbf{S}}.
\end{equation}
The regularizer is defined as
\begin{equation}
\mathcal{L}_{\mathrm{sep}}
=
\|\mathbf{C}-\mathbf{I}\|_F^2.
\label{eq:Lsep}
\end{equation}

This term encourages different latent dimensions to avoid collapsing into highly similar representations. In the present work, its role is intentionally weak: it does not replace the source-wise structural mechanism, but only provides an additional preference for low cross-correlation among latent components. Under the blind source separation interpretation, it may be viewed as a mild acceleration term that helps the optimization move more efficiently from ``good reconstruction but still partially entangled'' latent configurations toward cleaner source-wise separation.

\subsection{Learning Objective}

Combining the observation model and the general source-wise latent energy in \eqref{eq:general_structured_prior}, the posterior has the form
\begin{equation}
p(\mathbf{S}\mid \mathbf{Y},\theta)
\propto
p(\mathbf{Y}\mid \mathbf{S},\theta)\,
\prod_{j=1}^{n}\exp\!\big(-E_j(\mathbf{s}_j;\phi_j)\big).
\end{equation}

For clarity, the overall objective may be written explicitly as a function of the optimization variables and the observed data:
\begin{equation}
\begin{aligned}
\mathcal{L}(\mathbf{S},\theta,\{\phi_j\}_{j=1}^{n};\mathbf{Y})
&=
\mathcal{L}_{\mathrm{obs}}
+
\lambda_{\mathrm{str}}\sum_{j=1}^{n}E_j(\mathbf{s}_j;\phi_j) \\
&\quad +
\lambda_{\mathrm{sep}}\mathcal{L}_{\mathrm{sep}},
\end{aligned}
\label{eq:Ltotal_general_explicit}
\end{equation}
where $\mathbf{S}$ denotes the latent representation, $\theta$ denotes the parameters of the observation map, $\phi_j$ denotes the source-wise structural parameters associated with the $j$-th latent dimension, and $\mathbf{Y}$ is the observed multichannel signal. In the present GP-based instantiation, $\phi_j=\ell_j$ and \eqref{eq:Ltotal_general_explicit} becomes
\begin{equation}
\begin{aligned}
\mathcal{L}(\mathbf{S},\theta,\{\ell_j\}_{j=1}^{n};\mathbf{Y})
&=
\frac{1}{2\nu_y}
\|\mathbf{Y}-g_{\theta}(\mathbf{S})\|_F^2 \\
&\quad +
\lambda_{\mathrm{GP}}
\sum_{j=1}^{n}
\left(
\frac{1}{2}\mathbf{s}_j^\top\mathbf{K}_j^{-1}\mathbf{s}_j
+
\frac{1}{2}\log|\mathbf{K}_j|
\right) \\
&\quad +
\lambda_{\mathrm{sep}}
\|\mathbf{C}-\mathbf{I}\|_F^2.
\end{aligned}
\label{eq:Ltotal_expanded}
\end{equation}

The observed data $\mathbf{Y}$ are fixed, while the latent trajectories $\mathbf{S}$, the parameters $\theta$ of the observation-generation map, and the source-wise structural parameters $\{\phi_j\}_{j=1}^{n}$ are jointly optimized. Consequently, training is not merely a process in which a scalar loss decreases, but also a process in which different latent dimensions progressively adapt under different structural constraints and gradually converge toward different roles.

More specifically, during every epoch, the current latent trajectories $\mathbf{S}$ are mapped into reconstructed observations $\widehat{\mathbf{Y}}$ through the observation map; the observation-matching term aligns them with the observed multichannel data; the source-wise structural energy reshapes each latent dimension according to its own structural bias; and the separation penalty discourages overlap among latent factors. In the particular realization studied here, this structural bias is instantiated by GP-inspired covariance geometry. Repeating this process over epochs gradually transforms the random initial trajectories into structured latent dimensions whose parameters and representations no longer evolve in the same manner, but instead tend to converge toward different configurations.

This progressive differentiation is particularly important for interpretation. In the present paper, where the framework is examined in the concrete form of blind source separation, the convergence of different latent dimensions toward different structured configurations may be viewed as the process by which underlying source components become separated without direct source supervision. From a broader structured latent modeling viewpoint, the same phenomenon may also be interpreted as an unsupervised decoupling process, in which different latent dimensions acquire different structural identities through joint optimization.

\subsection{Optimization Characteristics}

Although the objective in \eqref{eq:Ltotal_expanded} is straightforward, its optimization behavior is not necessarily uniform across training. Since the latent trajectories are optimized directly together with the observation-map parameters and the source-wise structural parameters, the learning dynamics are governed by the joint interaction of data fitting, source-wise structural adaptation, and weak latent decorrelation. Consequently, the optimization process may be viewed as
\begin{equation}
\min_{\mathbf{S},\,\theta,\,\{\phi_j\}}
\mathcal{L}_{\mathrm{obs}}
+
\lambda_{\mathrm{str}}\sum_{j=1}^{n}E_j(\mathbf{s}_j;\phi_j)
+
\lambda_{\mathrm{sep}}\mathcal{L}_{\mathrm{sep}},
\label{eq:Lopt_characteristic}
\end{equation}
where the same latent representation must simultaneously satisfy observation matching and source-wise structural differentiation.

In practice, this often leads to a two-stage optimization pattern. The early stage is usually dominated by observation fitting, so that the latent representation is rapidly adjusted into a configuration that explains the observed mixtures reasonably well. The later stage is typically slower, because the model must further redistribute structural roles across latent dimensions under the source-wise energies, even when the observation mismatch has already become small. From the source separation viewpoint, this means that satisfactory reconstruction of the mixtures may be reached earlier than clean source-wise decoupling.

This effect becomes more visible when the observation map is highly flexible, since part of the fitting burden may then be absorbed by the mapping itself, reducing the pressure for the latent trajectories to settle immediately into distinct source-wise configurations. Accordingly, the convergence behavior of the proposed framework should be interpreted not only by the decrease of the overall loss, but also by the gradual differentiation of latent roles during training. More detailed empirical manifestations of this phenomenon will be shown in the case study section.

\begin{algorithm*}[t]
\scriptsize
\setlength{\algomargin}{0.55em}
\SetAlgoLined
\DontPrintSemicolon
\SetNlSkip{0.15em}
\SetInd{0.15em}{0.35em}
\linespread{0.92}\selectfont
\KwIn{Normalized observations $\widetilde{\mathbf{Y}}\in\mathbb{R}^{T\times m}$, normalized sample index $\mathbf{t}=[t_1,\dots,t_T]^\top$, source number $n$, weights $\nu_y,\lambda_{\mathrm{GP}},\lambda_{\mathrm{sep}}$, initialization scale $\sigma_{\mathrm{init}}$.}
\KwOut{Estimated sources $\widehat{\mathbf{S}}$, observation-map parameters $\widehat{\theta}$, source-wise length-scales $\{\widehat{\ell}_j\}_{j=1}^{n}$.}

Initialize $\mathbf{S}=\sigma_{\mathrm{init}}\boldsymbol{\Xi}$ with $\Xi_{ij}\sim\mathcal{N}(0,1)$, initialize $\theta$, initialize $\{\eta_j\}_{j=1}^{n}$, and set $\ell_j=\exp(\eta_j)$\;

\While{not converged}{
    $\widehat{\mathbf{Y}}=g_{\theta}(\mathbf{S})$\;
    
    $\mathcal{L}_{\mathrm{obs}}=\dfrac{1}{2\nu_y}\left\|\widetilde{\mathbf{Y}}-\widehat{\mathbf{Y}}\right\|_F^2$\;
    
    \For{$j=1$ \KwTo $n$}{
        $\mathbf{K}_j(i,r)=\sigma_f^2\exp\!\left(-\dfrac{(t_i-t_r)^2}{2\ell_j^2}\right)+\epsilon\delta_{ir}$\;
        
        Compute the Cholesky factorization $\mathbf{K}_j=\mathbf{L}_j\mathbf{L}_j^{\top}$ and obtain $\mathbf{a}_j=\mathbf{K}_j^{-1}\mathbf{s}_j$ via triangular solves\;
        
        $E_j=\dfrac{1}{2}\mathbf{s}_j^{\top}\mathbf{a}_j+\sum_{i=1}^{T}\log(\mathbf{L}_j)_{ii}$\;
    }
    
    $\mathcal{L}_{\mathrm{GP}}=\sum_{j=1}^{n}E_j$\;
    
    Normalize each column of $\mathbf{S}$ to obtain $\widetilde{\mathbf{S}}$, then set $\mathbf{C}=\dfrac{1}{T}\widetilde{\mathbf{S}}^{\top}\widetilde{\mathbf{S}}$\;
    
    $\mathcal{L}_{\mathrm{sep}}=\left\|\mathbf{C}-\mathbf{I}\right\|_F^2$\;
    
    $\mathcal{L}=\mathcal{L}_{\mathrm{obs}}+\lambda_{\mathrm{GP}}\mathcal{L}_{\mathrm{GP}}+\lambda_{\mathrm{sep}}\mathcal{L}_{\mathrm{sep}}$\;
    
    $\{\mathbf{S},\theta,\eta_1,\dots,\eta_n\}\leftarrow\mathrm{OptimizerStep}(\nabla\mathcal{L})$\;
    
}
Set $\widehat{\mathbf{S}}=\mathbf{S}$, $\widehat{\theta}=\theta$, and $\widehat{\ell}_j=\ell_j$ for $j=1,\dots,n$\;
\caption{Training procedure of the source-wise GP-inspired energy-based model.}
\label{alg:sourcewise_gp_ebm}
\end{algorithm*}

\section{Experimental Study}

This section evaluates the proposed source-wise GP-inspired energy-based model under both linear and nonlinear mixing settings. The experiments are designed not only to verify final separation accuracy, but also to examine the optimization behavior emphasized in the methodology, namely that source separation is achieved through the joint evolution of the latent trajectories, the observation map, and the source-wise GP parameters. In particular, special attention is paid to the role of the auxiliary separation regularizer, since the methodology already suggested that structured source-wise energies alone may be sufficient for eventual separation, while the regularizer mainly affects the convergence path rather than the basic feasibility of recovery.

\subsection{Experimental Setup}

A synthetic multichannel source separation task is considered, with three latent/source components defined over $T=1000$ indexed samples. The observed mixtures are standardized before training, and the sample index is normalized to $[0,1]$ for constructing the source-wise RBF covariance matrices. In all experiments, the latent matrix is directly initialized by Gaussian noise and optimized jointly with the observation map and the source-wise GP length-scales, in accordance with the formulation in Section~\ref{alg:sourcewise_gp_ebm}. 

Two observation settings are examined.

\begin{enumerate}
    \renewcommand{\labelenumi}{\alph{enumi})}
    \item \textbf{Linear case.} The observation map is a linear mixer without hidden layers or bias. This setting is used to examine the behavior of the proposed source-wise GP energy under a standard linear mixing assumption.
    \item \textbf{Nonlinear case.} The observation map is replaced by a row-wise multilayer perceptron with two hidden layers, allowing nonlinear mapping from latent components to observations.
\end{enumerate}

To evaluate source recovery, the estimated latent components are aligned to the true sources by permutation matching, and the absolute correlation is reported for monitoring. It is important to emphasize that this correlation is \emph{not} included in the training objective. It is used only as an external diagnostic to visualize how the latent trajectories progressively evolve toward the underlying sources during optimization. Therefore, the monitoring curves should be interpreted as evaluation signals rather than optimization targets.

\subsection{Linear Mixing Scenario Without Explicit Separation Regularization}

\begin{figure*}[t]
    \centering
    \includegraphics[width=0.92\textwidth]{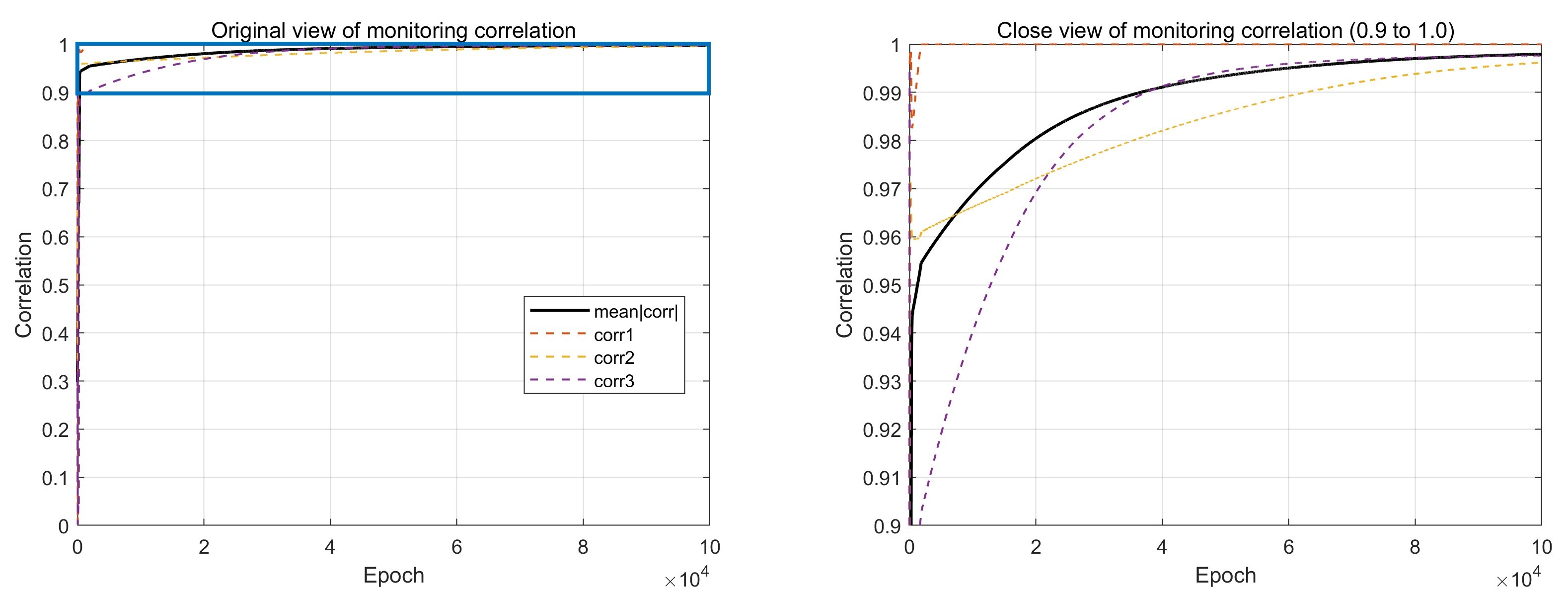}
    \caption{Linear-case monitoring correlation without the separation regularizer.}
    \label{fig:linear_no_sep_monitor}
\end{figure*}

We first consider the linear case without the explicit separation term, corresponding to the setting in Figure~\ref{fig:linear_no_sep_monitor}. This experiment is important because it connects the present model to the author's previous structured latent designs, such as SAHMM-VAE (\cite{wei2026sahmm}), PDGMM-VAE (\cite{wei2026pdgmmvae}), and StrADiff (\cite{wei2026stradiff}), where source-wise structural design alone was already sufficient to drive separation without introducing an additional decorrelation penalty.

The result shows that the proposed model can indeed recover the three sources accurately even without the separation term. This confirms that the essential separation mechanism still comes from the combination of observation matching and source-wise structured GP energy, rather than from an explicit correlation penalty. In other words, the source-wise structural design itself remains the principal mechanism that differentiates the latent dimensions and drives them toward distinct source-like roles.

However, Figure~\ref{fig:linear_no_sep_monitor} also reveals a characteristic optimization phenomenon. Although the monitoring correlation rises quickly at the beginning of training, the subsequent improvement becomes very slow, and the model spends a long range of epochs making only marginal gains. This behavior is fully consistent with the discussion in the methodology section: once the mixtures are already reconstructed reasonably well, the optimization enters a relatively flat late stage in which further refinement of the latent trajectories toward cleaner source-wise decoupling proceeds only gradually. Thus, the no-separation setting demonstrates that the model is capable of separation on its own, but that the late-stage convergence can be inefficient.

\subsection{Effect of the Separation Regularizer in the Linear Case}

\begin{figure*}[t]
    \centering
    \includegraphics[width=0.92\textwidth]{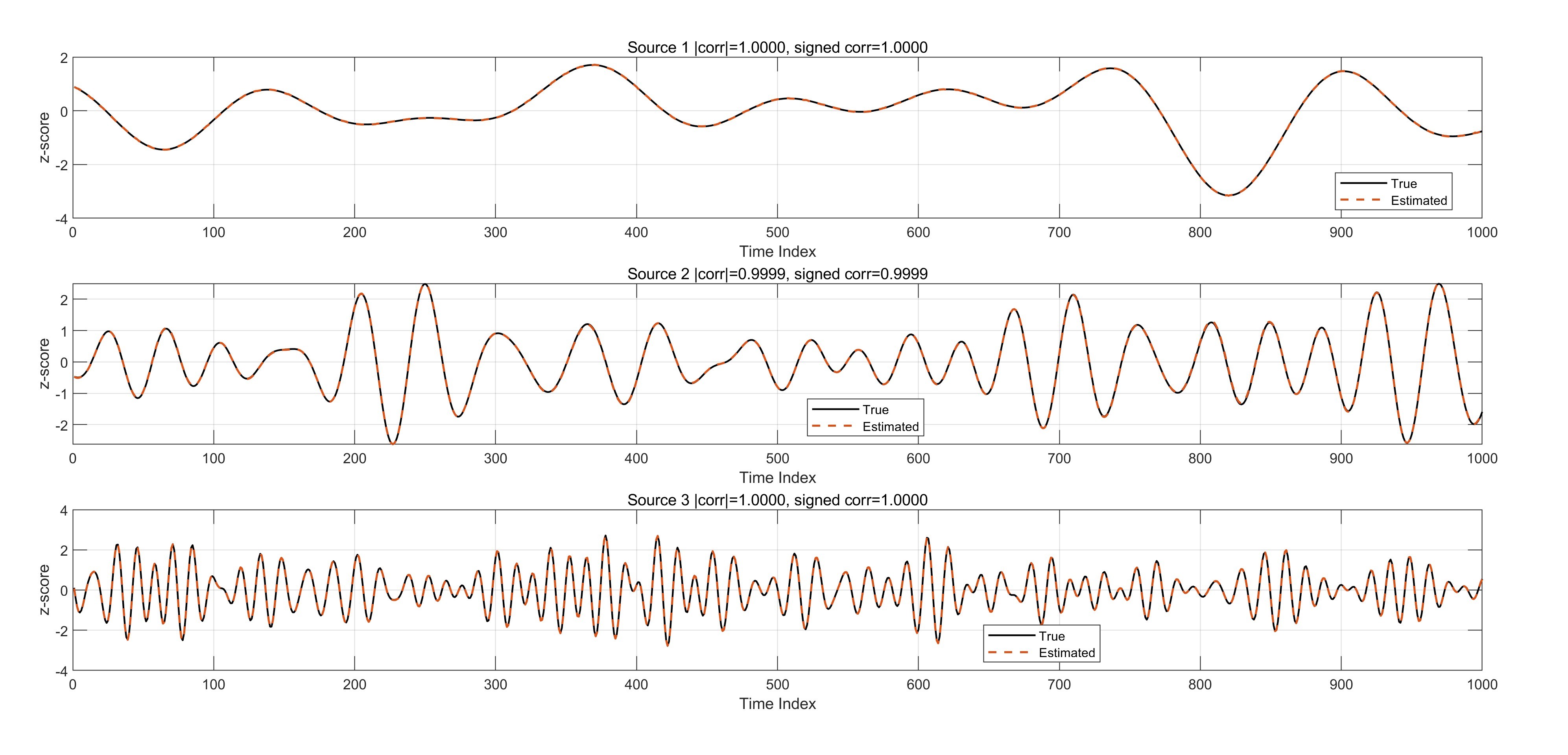}
    \caption{Linear-case matched source comparison with the separation regularizer.}
    \label{fig:linear_with_sep_sources}
\end{figure*}

\begin{figure*}[t]
    \centering
    \includegraphics[width=0.92\textwidth]{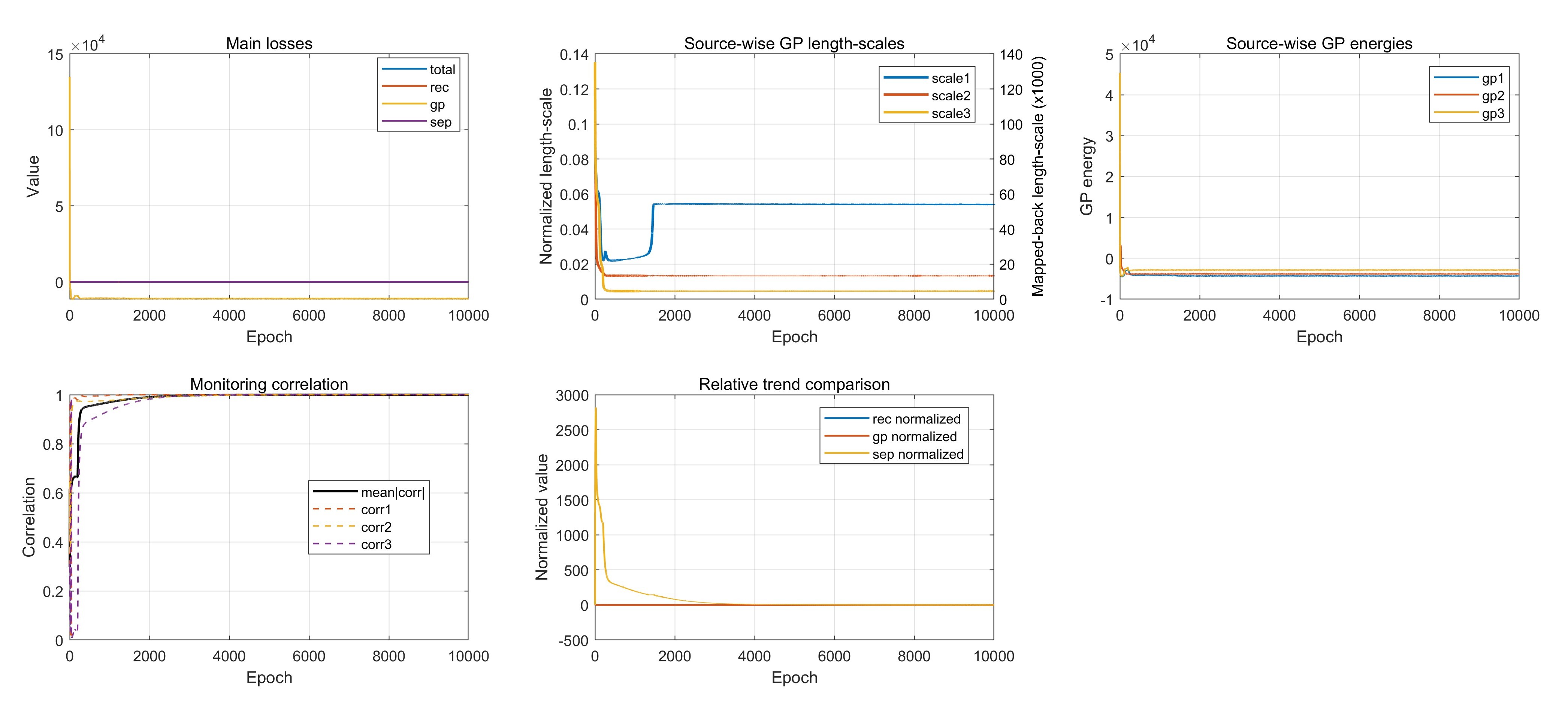}
    \caption{Training dynamics in the linear case with the separation regularizer, including losses, source-wise GP length-scales, GP energies, and monitoring correlation.}
    \label{fig:linear_with_sep_dynamics}
\end{figure*}

To address the slow late-stage refinement observed above, the auxiliary separation regularizer is activated in the linear case. The corresponding source recovery result is shown in Figure~\ref{fig:linear_with_sep_sources}, and the training dynamics are summarized in Figure~\ref{fig:linear_with_sep_dynamics}.

Figure~\ref{fig:linear_with_sep_sources} shows that, after adding the separation term, the recovered sources become almost indistinguishable from the ground truth. More importantly, the main improvement is not merely the final accuracy, but the optimization speed toward that accuracy. In Figure~\ref{fig:linear_with_sep_dynamics}, the monitoring-correlation subplot shows a much faster rise compared with the no-separation case, indicating that the latent components are pushed toward source-wise decoupling much earlier during training. This observation is consistent with the intended role of $\mathcal{L}_{\mathrm{sep}}$: it does not create the separation mechanism by itself, but reduces redundant overlap among latent dimensions and helps the optimizer leave the regime of ``good reconstruction but still partially entangled'' representations more quickly.

The remaining subplots in Figure~\ref{fig:linear_with_sep_dynamics} provide further insight into this process. The loss curves indicate stable joint optimization of the reconstruction term, GP energy, and separation term. The learned source-wise length-scales settle at clearly different values, suggesting that different latent dimensions indeed converge to different structural regimes rather than sharing one common temporal scale. This is precisely the source-wise structural effect that the model is designed to induce. At the same time, the GP-energy trajectories remain well behaved, showing that the acceleration in source recovery is not obtained by discarding the structured prior, but by coordinating it with an additional weak decorrelation preference.

An important point should again be emphasized here: the correlation curves in Figures~\ref{fig:linear_no_sep_monitor} and~\ref{fig:linear_with_sep_dynamics} are \emph{not} optimization objectives. They are computed only for monitoring against the known synthetic ground truth. Therefore, the faster rise of correlation after introducing $\mathcal{L}_{\mathrm{sep}}$ should be interpreted as evidence that the auxiliary regularizer improves the latent optimization path, not that the model is being directly trained to maximize correlation.

Overall, the linear experiments support two conclusions. First, the proposed source-wise GP-inspired energy is already capable of driving source separation without explicit separation loss, which is consistent with the behavior observed in the author's previous structured models. Second, in the present EBM-style formulation, adding a weak separation term substantially improves convergence efficiency, especially in the later stage where pure structured optimization becomes slow.

\subsection{Nonlinear Mixing Scenario}

\begin{figure*}[t]
    \centering
    \includegraphics[width=0.92\textwidth]{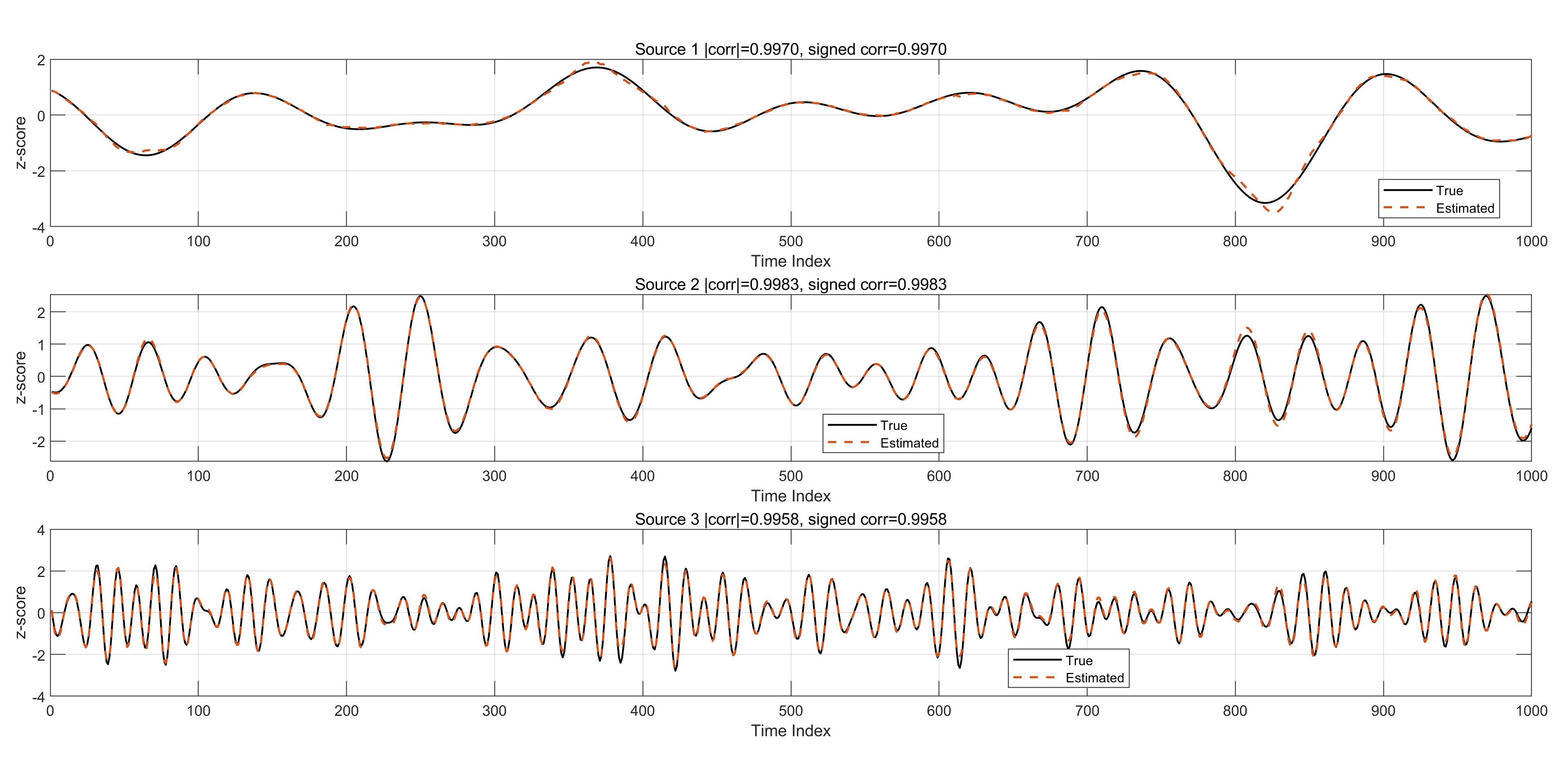}
    \caption{Nonlinear-case matched source comparison.}
    \label{fig:nonlinear_sources}
\end{figure*}

\begin{figure*}[t]
    \centering
    \includegraphics[width=0.92\textwidth]{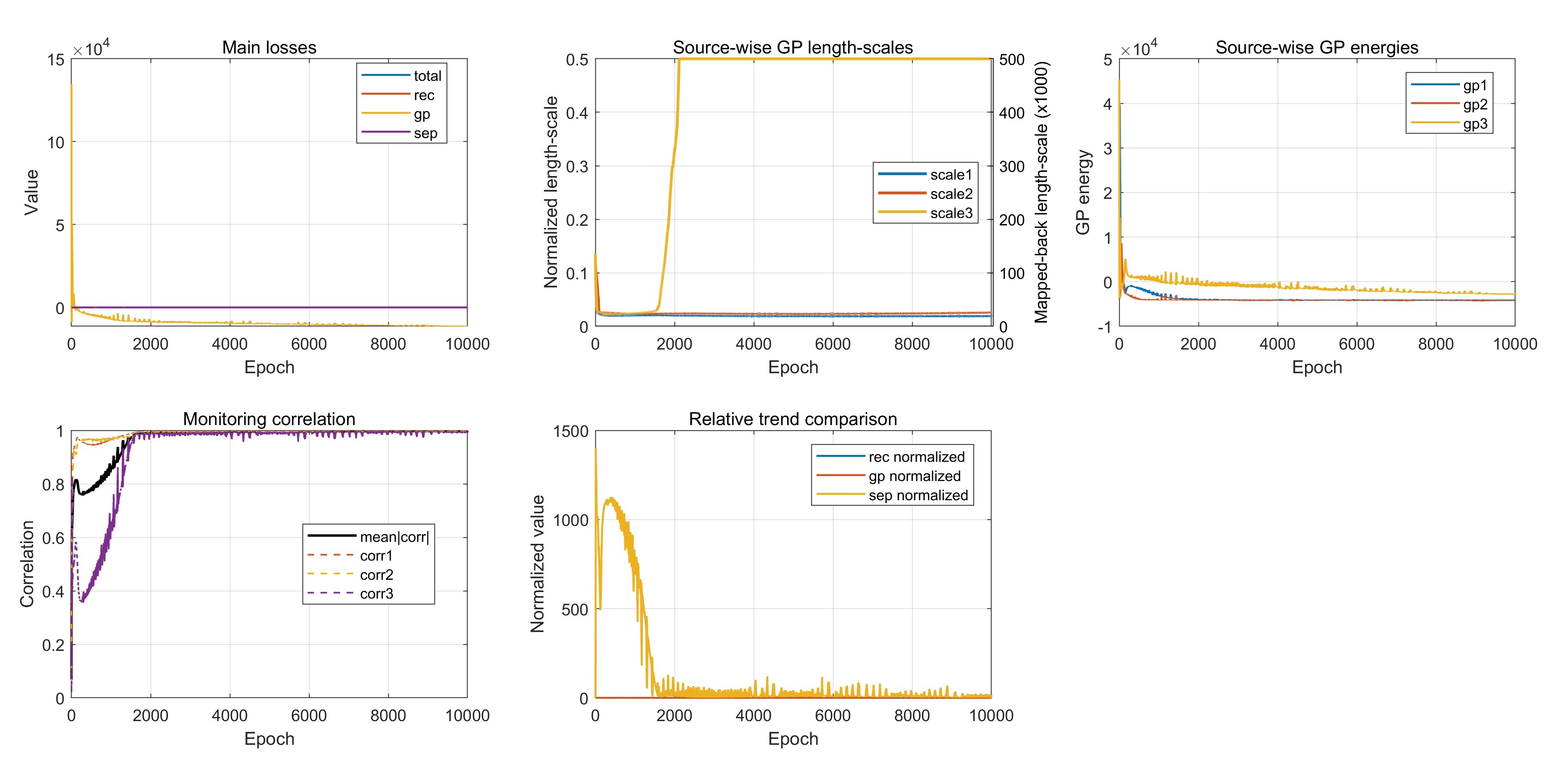}
    \caption{Training dynamics in the nonlinear case, showing losses, source-wise GP length-scales, GP energies, and monitoring correlation.}
    \label{fig:nonlinear_dynamics}
\end{figure*}

We next examine the nonlinear case, in which the linear mixer is replaced by a multilayer perceptron. The final matched source comparison is shown in Figure~\ref{fig:nonlinear_sources}, and the corresponding optimization behavior is summarized in Figure~\ref{fig:nonlinear_dynamics}.

Figure~\ref{fig:nonlinear_sources} shows that the proposed model still retains a substantial separation capability under nonlinear mixing. The recovered components remain highly consistent with the true sources, indicating that the source-wise GP-inspired latent structure is not limited to the linear case. This is an important result, because it suggests that the same structured latent principle can still function when the observation-generation mechanism becomes more flexible.

Nevertheless, the nonlinear case is visibly less stable than the linear one. Compared with the nearly perfect linear recovery, the source matching in Figure~\ref{fig:nonlinear_sources} is slightly weaker, and the training traces in Figure~\ref{fig:nonlinear_dynamics} show clearer fluctuations. In particular, the monitoring correlation no longer exhibits the same smooth and steadily improving behavior as in the linear setting. Instead, it oscillates during training, indicating that the increased flexibility of the nonlinear observation map makes the latent optimization more difficult. This is also consistent with the optimization analysis in the methodology: when the observation map is more expressive, part of the fitting burden can be absorbed by the map itself, reducing the immediate pressure for the latent dimensions to settle into cleanly differentiated source-wise configurations.

A further phenomenon appears in the learned length-scales. In Figure~\ref{fig:nonlinear_dynamics}, the third source-wise length-scale reaches the clamp boundary. This indicates that, for that latent dimension, the length-scale parameter does not converge to a well-internalized structural optimum, but is instead pushed to the imposed feasible range. From an optimization perspective, this suggests that the third latent component is struggling to establish a stable GP scale under the nonlinear mapping. Therefore, although the model still achieves reasonably strong source recovery, the nonlinear case does not exhibit the same level of robustness and stability observed in either the linear setting of the present method or several of the author's previous structured generative models.

In this sense, the nonlinear experiment should be interpreted as both a positive and a cautionary result. On the positive side, the method remains workable beyond linear mixing. On the cautionary side, the optimization becomes more fragile, the source-wise GP parameters are more difficult to stabilize, and the separation trajectory is less smooth. These observations suggest that, under nonlinear mixing, the interaction among latent trajectories, GP structural energies, and a flexible observation map becomes substantially more complicated than in the linear case.

\subsection{Discussion}

Taken together, the experiments reveal a rather clear empirical picture of the proposed framework.

First, the source-wise GP-inspired structural design is the main mechanism enabling separation. Even without the explicit separation term, the model can already recover the sources in the linear case, which is fully in line with the conceptual motivation of the methodology and with the author's earlier structured latent models.

Second, the separation regularizer plays an important practical role in the present EBM-style formulation. Its main contribution is not to make separation possible, but to accelerate source-wise differentiation and improve optimization efficiency. This is especially evident from the comparison between Figures~\ref{fig:linear_no_sep_monitor} and~\ref{fig:linear_with_sep_dynamics}, where the monitored source correlations improve much faster after the regularizer is added, even though correlation itself is never part of the training loss.

Third, the linear and nonlinear cases differ not only in final accuracy but also in optimization character. The linear case is remarkably stable and leads to nearly exact source recovery. The nonlinear case still works, but exhibits visibly larger fluctuations and less stable length-scale adaptation. In particular, the clamped behavior of one learned length-scale suggests that nonlinear flexibility can weaken the stability of source-wise structural convergence.

Overall, these results support the central claim of this paper: directly optimized latent trajectories, when equipped with source-wise GP-inspired energies, can evolve into distinct source-like components through joint optimization with the observation map. At the same time, the experiments also clarify an important practical point: for this class of source-wise EBM formulations, auxiliary separation regularization can be highly beneficial for accelerating convergence, whereas nonlinear observation maps introduce additional instability that deserves further investigation in future work.

\section{Conclusion}

This paper proposed StrEBM, a structured latent energy-based model for source-wise structured representation learning, and examined it in the concrete setting of blind source separation. The central idea is to assign each latent dimension its own learnable structural energy, so that latent components are not merely regularized by a single shared constraint, but are gradually driven toward different source-like roles through joint optimization with the observation map. In the present study, this source-wise design was instantiated using GP-inspired energies with learnable length-scales, providing a concrete realization of a more general structured latent EBM framework.

The experiments on synthetic multichannel signals showed that the proposed framework is able to recover underlying source components effectively in both linear and nonlinear mixing settings. In the linear case, the model exhibited strong separation performance and stable optimization behavior, confirming that source-wise structural energy itself can already serve as the main mechanism driving latent differentiation. The results also showed that the auxiliary separation regularizer is not the origin of separability, but a practically useful term for accelerating late-stage convergence and reducing residual overlap among latent dimensions.

At the same time, the case study also clarified an important limitation of the current formulation. Although the nonlinear setting remained workable, its optimization was visibly less stable than that of the linear case, with stronger fluctuations in the monitoring curves and less robust convergence of the source-wise structural parameters. A plausible reason is that, once the observation map becomes sufficiently flexible, the current formulation may no longer provide enough effective constraints on the interaction among latent trajectories, source-wise energies, and the nonlinear generator. As a result, part of the data fitting can be absorbed by the nonlinear mapping itself, weakening the pressure for the latent dimensions to settle into stable and clearly differentiated structural roles.

Therefore, beyond establishing an initial proof of concept, this work also points to an important future direction. In particular, the nonlinear case suggests that stronger or more principled forms of constraint may be necessary to stabilize source-wise latent differentiation under expressive observation mappings. One possible direction is to introduce variational inference or related constrained inference mechanisms, so that the latent optimization is no longer driven only by direct energy shaping and reconstruction matching, but is additionally guided by a more explicit posterior regularization principle. More generally, other forms of structural constraints, parameter-sharing restrictions, or source-wise probabilistic formulations may also be explored to improve robustness in nonlinear settings.

Overall, the present work should be viewed as a first step toward a broader source-wise structured latent EBM framework. The GP-based instantiation studied here provides an empirically verifiable example showing that direct latent optimization under source-wise energies can promote source-like decoupling. At the same time, the observed nonlinear instability indicates that richer constraint mechanisms and more robust optimization strategies will be essential for extending this idea toward more general identifiable, interpretable, and nonlinear structured latent learning problems.

\bibliography{ref}

\end{document}